\documentclass[conference]{IEEEtran}
\IEEEoverridecommandlockouts

\usepackage{cite}
\usepackage{amsmath,amssymb,amsfonts}
\usepackage{algorithmic}
\usepackage{graphicx}
\usepackage{textcomp}
\usepackage[table,dvipsnames]{xcolor}
\def\BibTeX{{\rm B\kern-.05em{\sc i\kern-.025em b}\kern-.08em
    T\kern-.1667em\lower.7ex\hbox{E}\kern-.125emX}}

\usepackage{colortbl} 
\definecolor{lightblue}{rgb}{0.9, 0.92, 1.0}
\definecolor{figure_orange}{RGB}{194,81,21}
\definecolor{figure_green}{RGB}{59,127,35}
\definecolor{figure_blue}{RGB}{32,92,150}
\colorlet{tablerowcolor}{lightblue}
\newcommand{\coloredmidrule}{\arrayrulecolor{white}\specialrule{\aboverulesep}{0pt}{0pt}%
            \arrayrulecolor{black}\specialrule{\lightrulewidth}{0pt}{0pt}%
            \arrayrulecolor{tablerowcolor}\specialrule{\belowrulesep}{0pt}{0pt}%
            \arrayrulecolor{black}}

\newcommand{\coloredbottomrule}{\arrayrulecolor{tablerowcolor}\specialrule{\aboverulesep}{0pt}{0pt}%
            \arrayrulecolor{black}\specialrule{\heavyrulewidth}{0pt}{\belowbottomsep}}%
\usepackage{makecell}
\usepackage{booktabs}       
\usepackage[pagebackref]{hyperref}
\usepackage{url}
\hypersetup{
    colorlinks=true,
    citecolor =cyan,
    linkcolor=magenta,
    urlcolor=blue,
}
            
\begin{document}

\title{Get Large Language Models Ready to Speak:\\A Late-fusion Approach for Speech Generation}
\author{%
  \IEEEauthorblockN{Maohao Shen\IEEEauthorrefmark{1}\thanks{*Work done during internship at AI@Meta. Audio samples are included in our project page: \url{https://maohaos2.github.io/TTS-Llama-MoLE-Llama/}.},
                    Shun Zhang\IEEEauthorrefmark{2},
                    Jilong Wu\IEEEauthorrefmark{2},
                    Zhiping Xiu\IEEEauthorrefmark{2},
                    Ehab AlBadawy\IEEEauthorrefmark{2},
                    Yiting Lu\IEEEauthorrefmark{2},
                    Mike Seltzer\IEEEauthorrefmark{2},
                    Qing He\IEEEauthorrefmark{2}
                }
\IEEEauthorblockA{\IEEEauthorrefmark{1}%
                   Massachusetts Institute of Technology}
\IEEEauthorblockA{\IEEEauthorrefmark{2}%
                    AI at Meta}
}

\maketitle

\begin{abstract}
Large language models (LLMs) have revolutionized natural language processing (NLP) with impressive performance across various text-based tasks. However, the extension of text-dominant LLMs to with speech generation tasks remains under-explored. In this work, we introduce a text-to-speech (TTS) system powered by a fine-tuned Llama model, named TTS-Llama, that achieves state-of-the-art speech synthesis performance. Building on TTS-Llama, we further propose MoLE-Llama, a text-and-speech multimodal LLM developed through purely late-fusion parameter-efficient fine-tuning (PEFT) and a mixture-of-expert architecture. Extensive empirical results demonstrate MoLE-Llama’s competitive performance on both text-only question-answering (QA) and TTS tasks, mitigating catastrophic forgetting issue in either modality. Finally, we further explore MoLE-Llama in text-in-speech-out QA tasks, demonstrating its great potential as a multimodal dialog system capable of speech generation.
\end{abstract}


\section{Introduction}
Large language models (LLMs) \cite{dubey2024llama, achiam2023gpt} have achieved remarkable success across various natural language processing (NLP) tasks, such as question answering~\cite{hendrycks2020measuring}, machine translation \cite{zhu2020incorporating}, and commonsense reasoning \cite{zellers2018swag}. More recent progress has emerged to extend LLMs to various modalities beyond text~ \cite{wu2023next, tang2024any, ye2024x}, among which the speech output modality is of great interest. 

The development of LLMs (or general artificial intelligent systems) with speech generation capability is closely related to the classic research area in text-to-speech (TTS) systems. Leveraging discrete speech representations \cite{defossez2022high}, recent approaches \cite{zhang2023speak, wang2023neural, kharitonov2023speak} reformulate TTS based on a language modeling task akin to LLMs. However, these methods typically focus solely on the TTS task and require training a speech language model from scratch, which tend to be demanding on both computation and speech data.

Another recent line of research is to integrate speech modality into LLMs pretrained only on text. This leverages the prowess of LLMs in text-based tasks and the premise that speech and text tasks potentially share synergy. While most existing efforts focus on extending pretrained text-based LLMs to handle speech input, such as speech understanding tasks~\cite{fathullah2024prompting, wu2023decoder, tang2023salmonn}, enabling LLMs in speech generation remains a challenging and under-explored area \cite{hao2023boosting}.

In this work, we aim to endow text-based LLMs with speech generation capabilities through purely parameter-efficient fine-tuning (PEFT)~\cite{hu2021lora} in lieu of full pretraining or fine-tuning. Moreover, we demonstrate the effectiveness of a mixture-of-experts architecture in both text and speech generation without compromising either modality. The three main contributions of this work are summarized as follows.

\begin{enumerate}
    \item In Section \ref{sec:tts-llama}, we present TTS-Llama, a TTS system based on a Llama 3-8B-Instruct model fined-tuned with LoRA. To the best of our knowledge, TTS-Llama is the first TTS system to achieve state-of-the-art performance by only PEFT fine-tuning a text-based LLM.
    \item In Section \ref{sec:mole-llama}, we introduce MoLE-Llama, a text-speech multimodal LLM capable of text-only question answering (QA) and TTS synthesis, as well as (with a chain-of-modality technique) text-in-speech-out QA. To the best of our knowledge, MoLE-Llama is the first text-speech multimodal LLM achieved through purely late fusion, specifically PEFT fine tuning as opposed to additional pretraining or full fine-tuning.
    \item In Section \ref{sec:exp}, we demonstrate that MoLE-Llama retains text capabilities of the original text-only LLMs after introducing the speech output modality, addressing a common concern in existing multimodal LLMs.
\end{enumerate}
\section{Related Work}
\subsection{Text-to-Speech}
Traditional TTS systems \cite{shen2018natural, li2019neural, ren2019fastspeech} typically model speech generation as a continuous signal regression task of generating mel spectrograms that are then synthesized into speech waveforms by a vocoder. More recent research~\cite{zhang2023speak, wang2023neural, kharitonov2023speak, borsos2023audiolm} have adopted a language modeling architecture and discrete speech token representation. For instance, \cite{zhang2023speak} and \cite{wang2023neural} introduce neural codec language models to generate discrete acoustic tokens encoded from Residual Vector Quantization (RVQ)-based models \cite{defossez2022high}. For speech synthesis, \cite{borsos2023audiolm} explores the use of two types of speech tokens, i.e., high-level semantic tokens and low-level acoustic tokens. Followed by this paradigm, \cite{kharitonov2023speak} adapts the two types of speech tokens for TTS, demonstrating enhanced naturalness and acoustic quality. \cite{peng2024voicecraft} further address both speech editing and TTS tasks.

However, most of the aforementioned methods train a speech language model from scratch and focus solely on TTS. Augmenting pretrained text LLMs with speech generation capability remains largely under-explored. Recently, \cite{hao2023boosting} integrates pretrained text LLM and speech LM~\cite{wang2023neural}, but is not able to achieve promising speech generation performance with fine-tuning alone.

\subsection{Text-Speech Multimodal Language Model}
Representing speech as discrete tokens~\cite{lakhotia2021generative} enables the joint modeling of text and speech modalities in a single text-and-speech language model. For example, \cite{wang2023viola, chen2023lauragpt, rubenstein2023audiopalm} each demonstrates a single speech language model capable of various classic speech-related tasks including speech-to-speech/text translation, speech recognition and text-to-speech synthesis. \cite{hassid2024textually} and \cite{maiti2024voxtlm} show further benefits of training speech language models initialized with pretrained text LLMs. Although these models encapsulate multiple speech-related tasks, they lack any conversation or reasoning capabilities possessed by state-of-the-art text LLMs.

Recent works also try to inject speech dialog and continuation capability into existing pretrained text LLMs. 
To that end, \cite{zhang2023speechgpt} and \cite{nachmani2023spoken} introduce a ``chain-of-modality'' technique, where the model generates text as an extra step before speech generation in the same decoding pass. \cite{chou2023toward} and \cite{nguyen2024spirit} utilize interleaved text and speech data in training LLMs to improve text-speech alignment.  However, these approaches require computationally expensive full-scale pretraining or fine-tuning of the foundational language model. Moreover, the loss of pretrained LLM's text capabilities has been an overlooked concern except in \cite{nguyen2024spirit}, which shows performance regression in text QA after introducing speech generation capability.

\section{TTS-Llama: A TTS System powered by Fine-tuned Text Llama Model} \label{sec:tts-llama}

\begin{figure}[!t]
    \centering
    \includegraphics[width=1.0\columnwidth]{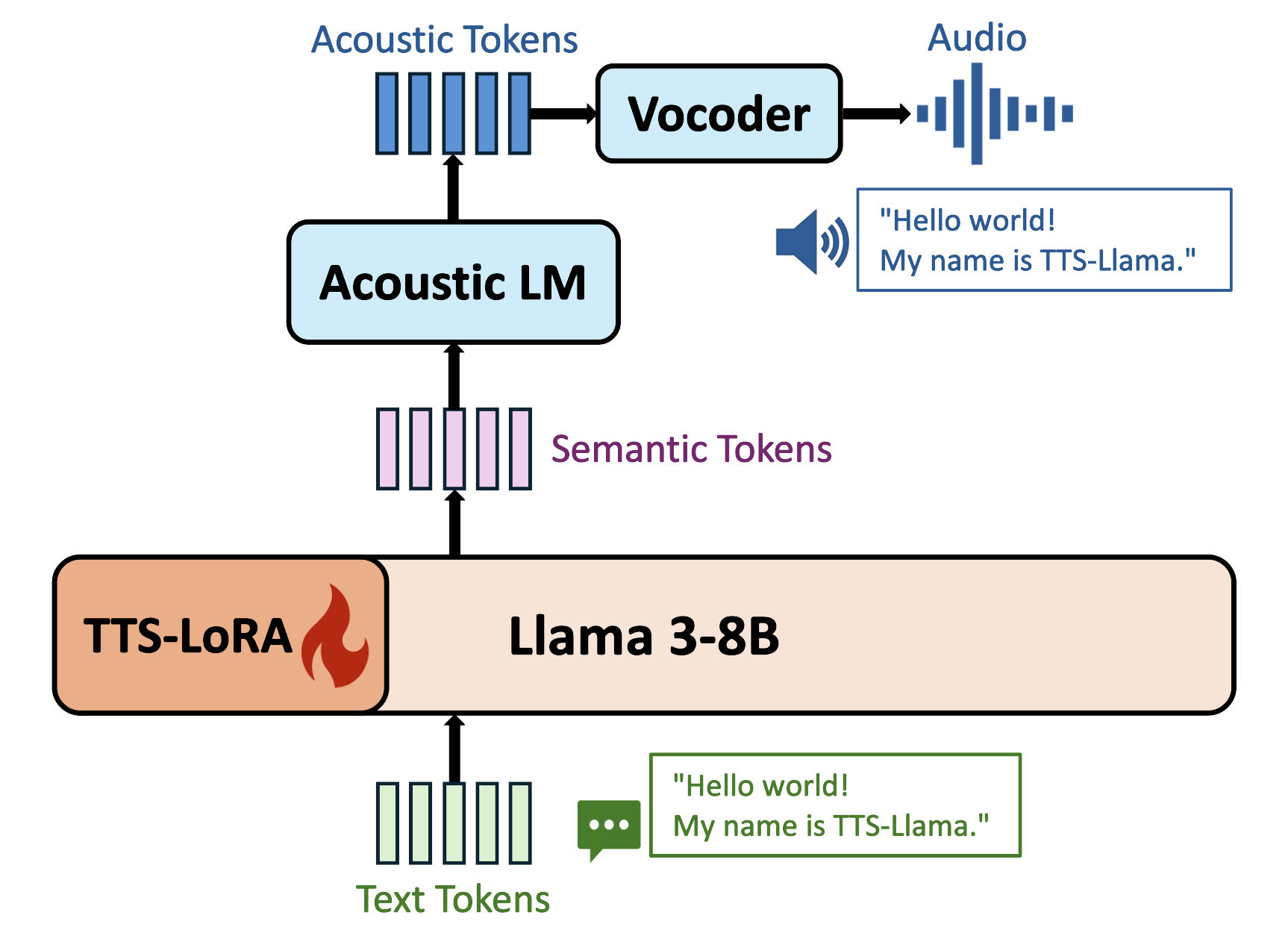}
    \vspace{-1em}
\caption{\textbf{Overview of TTS-Llama.} The core engine of the TTS system is the fine-tuned Llama model, which extracts high-level semantic information from the input text. An acoustic model, conditioned on this semantic information, further extracts low-level acoustic features for speech synthesis.}
\label{figure:tts-llama}
\vspace{-1em}
\end{figure}

To achieve high stability and naturalness in TTS, our proposed TTS-Llama model tackles the TTS task through a two-step speech token generation process. First, the fine-tuned Llama model processes raw text input to generate high-level semantic tokens that contains both semantic and prosody information. Then, an acoustic language model (LM) translates these semantic tokens into low-level acoustic features. Finally, a neural vocoder synthesizes the audio waveform from these acoustic features. To construct the training targets for the Llama model and the acoustic LM, we use two tokenizers to extract semantic and acoustic information, projecting them into a quantized latent space to produce discrete tokens. The overall design of TTS-Llama is illustrated in Figure~\ref{figure:tts-llama}, and we describe each model component below.

\paragraph{Fine-tuned Llama Model} \label{subsec:finetuned-llama}
The Llama model is the core engine of the proposed TTS system, as the semantic tokens carry rich information that directly impacts audio generation. Leveraging a pretrained Llama model rather than training a model from scratch offers two key benefits. First, it is computationally efficient by leveraging Llama model's prior knowledge in understanding text. Second, incorporating a pretrained Llama model allows integration into tasks beyond speech (e.g., text and vision tasks), towards the development of a multimodal system. The fine-tuning procedure is straightforward. Specifically, we expand the vocabulary and prediction head of the pre-trained Llama3-8B-Instruct model \cite{dubey2024llama} to accommodate the generation of semantic tokens. We then fine-tune the Llama model using PEFT approach LoRA \cite{hu2021lora}. The trainable components include the input embedding layers, the prediction head, and the injected LoRA adapters.

\paragraph{Tokenizers}
The semantic tokenizer provides a high-level discretized representation of audio that removes low-level redundant acoustic details while preserving sufficient information to accurately reconstruct the original audio. Our semantic tokenizer design is based on \cite{zhang2023google}, utilizing a quantizer \cite{chiu2022self} to extract 4,096 discrete semantic tokens. The acoustic tokenizer is implemented as a convolutional autoencoder. It employs a residual vector quantizer (RVQ) with a multi-codebook bottleneck \cite{defossez2022high} to extract discrete acoustic tokens.

\paragraph{Acoustic LM and Vocoder}
The acoustic LM is an autoregressive transformer-based decoder inspired by MusicGen \cite{copet2024simple}. Unlike the text conditioning used in \cite{copet2024simple}, our acoustic model is conditioned on semantic tokens, which it uses as inputs to generate multi-codebook acoustic tokens. The acoustic LM is a lightweight model compared to Llama with fewer than 0.2B parameters. Finally, we re-utilize the decoder from the acoustic tokenizer \cite{chiu2022self} as the vocoder to convert the sequence of acoustic tokens back into a waveform.

\section{MoLE-Llama: A Text-Speech Multimodal LLM via Mixture-of-Lora Experts} \label{sec:mole-llama}
\begin{figure*}[!t]
    \centering
    \includegraphics[width=1.75\columnwidth]{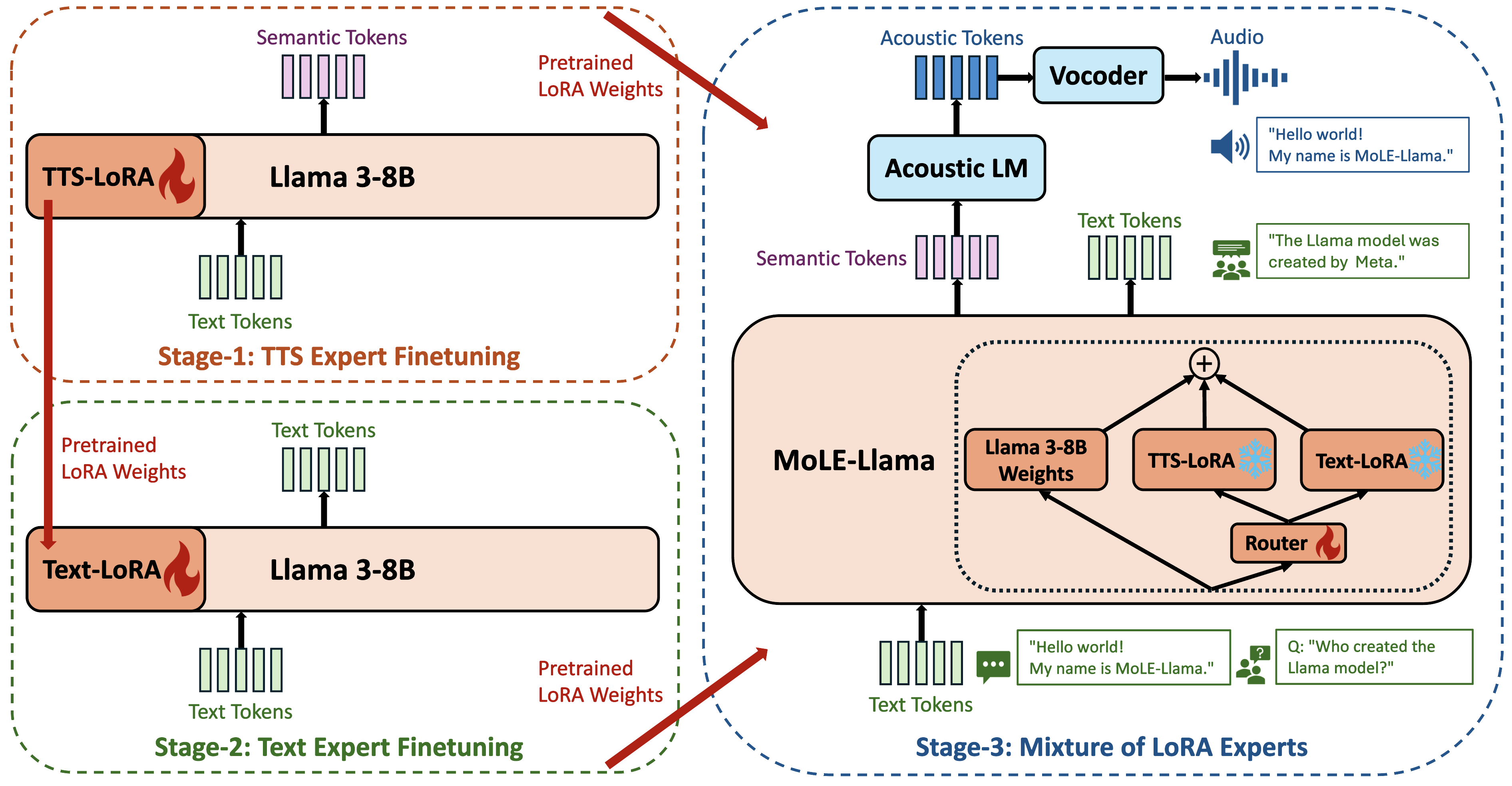}
    \vspace{-0.5em}
\caption{\textbf{Overview of MoLE-Llama.} MoLE-Llama is trained using a late-fusion approach consisting of three stages: \textcolor{figure_orange}{\textbf{Stage-1:}} Inject speech modality by fine-tuning a text-based Llama3-8B model for the TTS task; \textcolor{figure_green}{\textbf{Stage-2:}} Preserve model’s text capabilities by continuously fine-tuning the LoRA adapter using text instruct-tuning data; \textcolor{figure_blue}{\textbf{Stage-3:}} Unify the text and speech LoRA experts into a single multimodal LLM using the mixture-of-LoRA experts technique. MoLE-Llama can be extended to address additional tasks, such as speech QA, by training an extra speech QA LoRA expert during Stage-2 (see Section~\ref{subsec:speech-qa}).}
\label{figure:mole-llama}
\vspace{-1em}
\end{figure*}

While TTS-Llama explores the potential of PEFT fine-tuning pretrained text LLM for the TTS task, to produce a genuinely text-speech multimodal LLM requires retaining the original text capabilities. However, the MMLU evaluation of TTS-Llama shown in Table~\ref{table:text-qa} shows catastrophic forgetting in the text QA performance.

To effectively address this issue, we propose a late-fusion text-speech multimodal LLM, dubbed as MoLE-Llama, that utilizes a mixture-of-LoRA experts technique \cite{buehler2024x}. The key idea is to use text and speech experts trained separately to handle tasks in their respective modalities, and to eventually merge these dedicated experts into a unified multimodal LLM 

\subsection{Three-stage Training Procedure}
We propose a three-stage procedure for training MoLE-Llama, illustrated in Figure \ref{figure:mole-llama} and detailed as follows.

\paragraph{Stage-1: TTS Expert Fine-tuning}
In stage-1, we fine-tune the text Llama model following the same approach as TTS-Llama, as outlined in Section~\ref{sec:tts-llama}. This stage aims to inject speech modalities into the text Llama model, allowing it to process both text and semantic tokens while aligning the two modalities within the same embedding space.

\paragraph{Stage-2: Text Expert Fine-tuning}
Since stage-1 may negatively impact the Llama model’s text capabilities, stage 2 is designed to restore these text capabilities after speech modality injection. Specifically, we freeze the re-trained input embedding and prediction head from stage-1 and continuously fine-tune the LoRA adapter on the text QA task.

\paragraph{Stage-3: Mixture-of-LoRA Experts}
In stage-3, we unify the TTS expert and text expert into a single multimodal LLM. When given an input, MoLE-Llama intelligently selects the modality-aware expert to handle the task. In this stage, we only train the mixture-of-LoRA experts router \cite{buehler2024x}, keeping all other parameters frozen. The entire system is trained end-to-end using a combination of text QA and TTS data, allowing the router to flexibly adapt to both tasks.

\subsection{One Step Towards Speech QA} \label{subsec:speech-qa}
While the primary focus of this work is the TTS task, we demonstrate that MoLE-Llama has the potential to achieve the more ambitious goal of responding to user requests in both text and speech modalities. To take a step towards this goal, we further explore the text-in-speech-out speech QA task using the chain-of-modality instruct-tuning technique \cite{zhang2023speechgpt}. In stage-2, we further create a speech QA expert by fine-tuning the Llama model to generate a text tokens followed by its semantic tokens. In stage-3, we integrate the text QA and speech QA experts into a unified multimodal LLM.
\section{Experiments} \label{sec:exp}

\begin{table}[!t]
  \begin{center}
  \scriptsize
  \caption{\textbf{Zero-shot TTS MOS Score ($\uparrow$).} }
  \vspace{-1em}
  \begin{tabular}{ccc}
    \toprule
    \textbf{Methods} & \textbf{Human Likeliness} & \textbf{Audio Quality} \\
    \midrule
    Ground Truth & $3.41_{\pm0.13}$ & $3.96_{\pm0.12}$ \\
    \midrule
    Your TTS \cite{casanova2022yourtts} & $1.92_{\pm0.11}$ & $2.68_{\pm0.11}$ \\
    Voice Craft \cite{peng2024voicecraft} & $2.85_{\pm0.13}$ & $3.17_{\pm0.11}$ \\
    \coloredmidrule
    \rowcolor{lightblue}
    \textbf{TTS-Llama} & $3.07_{\pm0.10}$ & $3.47_{\pm0.15}$\\
    \rowcolor{lightblue}
    \textbf{MoLE-Llama (TTS+Text QA)} & $3.02_{\pm0.09}$ & $3.38_{\pm0.14}$\\
    \coloredbottomrule
  \end{tabular}
  \label{table:mos-general}
  \end{center}
  \vspace{-1.5em}
\end{table}

\begin{table}[!t]
  \begin{center}
  \scriptsize
  \caption{\textbf{Text Normalization TTS MOS Score ($\uparrow$).} }
  \vspace{-1em}
  \begin{tabular}{ccc}
    \toprule
    \textbf{Methods} & \textbf{Human Likeness} & \textbf{Audio Quality} \\
    \midrule
    Your TTS \cite{casanova2022yourtts} & $1.63_{\pm0.10}$ & $2.41_{\pm0.10}$ \\
    Voice Craft \cite{peng2024voicecraft} & $2.30_{\pm0.14}$ & $3.06_{\pm0.13}$ \\
    \coloredmidrule
    \rowcolor{lightblue}
    \textbf{TTS-Llama} & $2.74_{\pm0.11}$ & $3.40_{\pm0.18}$\\
    \rowcolor{lightblue}
    \textbf{MoLE-Llama (TTS+Text QA)} & $2.73_{\pm0.14}$ & $3.43_{\pm0.17}$\\
    \coloredbottomrule
  \end{tabular}
  \label{table:mos-tn}
  \end{center}
  \vspace{-1.5em}
\end{table}

\subsection{Settings} \label{subsec:setting}
\paragraph{Datasets}
Our speech training data consists of a mixture of open-source and in-house speech data, with the majority of the open-source data sourced from LibriHeavy~\cite{kang2024libriheavy}, totaling 50K hours of speech paired with text transcripts. To improve the model’s TTS performance on the text normalization (TN) task, we augment the main training dataset with about 60K internally curated sentences that are rich in various written-form text that requires text normalization sentences, as well as their corresponding synthetic speech generated by an internal TTS system.  For the text QA task, we use a subset of in-house Llama3 supervised fine-tuning data, consisting of 2M QA pairs. For the speech QA task, we filter out coding and math-heavy QA data, generating synthetic speech for the text responses, resulting in about 1M samples of of text instructions, text and speech responses.

\paragraph{Model Parameters}
The LoRA adapters are configured with a rank of 128 and alpha of 64 in all the experiments. The input embedding layer and prediction head of the Llama model are extended to accommodate the speech vocabulary of 4096 semantic tokens. This configuration results in 1.4B trainable parameters in MoLE-Llama Stage-1 (i.e. TTS-Llama) and 0.3B in Stage-2. The router used in Stage-3 is a lightweight MLP-based classifier, reducing the trainable parameters to 30M.

\paragraph{Training Details}
TTS-Llama and MoLE-Llama follow a supervised fine-tuning approach. We use the same prompt format as Llama Instruct fine-tuning \cite{dubey2024llama}, with different system prompts guiding the Llama model to adapt to specific tasks. The system prompts used are: ``Transform the input written-form English text into non-language tokens that represent the corresponding speech in audio'' for the TTS task, ``You are a helpful AI assistant'' for the text QA task, and ``Answer the input text questions using non-language tokens that represent the corresponding speech'' for the speech QA task. The models are optimized using the AdamW optimizer with an initial learning rate of $3\times 10^{-4}$ (except for $10^{-4}$ in MoLE-Llama Stage-3), optimizer parameters $\beta_1 = 0.9$ and $\beta_2 = 0.98$, and a cosine learning rate scheduler. Training is conducted on 8 H100 GPUs with a sequence length of 2048, a batch size of 256, and a maximum of 20K training steps.

\paragraph{Evaluation Metrics}
We conduct two separate evaluations to comprehensively evaluate the TTS performance of different methods. First, we evaluate the models on a zero-shot TTS task with randomly selected 50 samples from the test dataset. The TTS system synthesizes speech for target text transcripts given a voice prompt and its corresponding transcript. Second, we evaluate the robustness of TTS systems on text normalization task. We collect 25 test samples that requires text normalization, containing at least two special characters, such as digits, ZIP codes, and currency symbols.

For both evaluations, we conduct Mean Opinion Score (MOS) tests across two dimensions: ``human likeness'' which measures naturalness and correctness, and ``audio quality'' which assesses the clarity of the generated speech. Each audio sample receives 10 ratings, and the average score is reported with a $95\%$ confidence interval.

To evaluate the text capabilities of TTS-Llama and MoLE-Llama, we report the models' text accuracy on three widely used text QA benchmark datasets to examine the reasoning abilities of text LLMs: MMLU (5-shot)~\cite{hendrycks2020measuring}, GPQA (zero-shot)~\cite{rein2023gpqa}, and ARC Challenge (zero-shot) \cite{clark2018think}.

\paragraph{Baseline Methods}
For the TTS task, we compare our proposed methods, TTS-Llama and MoLE-Llama, against state-of-the-art TTS systems: YourTTS~\cite{casanova2022yourtts} and VoiceCraft \cite{peng2024voicecraft}. For the text QA task, we mainly compare our models with the open-source Llama3-8B-Instruct model checkpoint to evaluate how well text capabilities are retained. For additional reference, we include the MMLU results of SPIRIT-LM, a SOTA text-speech multimodal LLM~\cite{nguyen2024spirit}.

\subsection{Results} \label{subsec:results}
We present the results for three models: (1) TTS-LLaMA, (2) MoLE-LLaMA (TTS + Text QA), and (3) MoLE-LLaMA (Speech QA + Text QA). Speech QA results are primarily showcased with audio samples on the published project page.

\paragraph{TTS Task}
The MOS test results for the zero-shot TTS task and the text normalization (TN) TTS task are shown in Table~\ref{table:mos-general} and Table~\ref{table:mos-tn}, respectively. TTS-Llama outperforms all baselines in both tasks. Notably, while most TTS systems require a frontend module to handle the TN task, TTS-Llama achieve competitive results without such module. Additionally, the multimodal MoLE-Llama (TTS+text QA) performs comparably to TTS-Llama on the TTS task while offering additional text QA capability as evaluated below.

\paragraph{Text QA Task} \label{subsec:exp-text-qa}
\begin{table}[!t]
  \begin{center}
  \scriptsize
  \caption{\textbf{Text QA Test Accuracy ($\uparrow$).}}
  \vspace{-1em}
  \begin{tabular}{cccc}
    \toprule
    \textbf{Methods} & \textbf{MMLU} & \textbf{GPQA} & \textbf{ARC Challenge} \\
    \midrule
    Llama3-8B-Instruct \cite{dubey2024llama} & $67.1\%$ & $31.9\%$ & $79.5\%$ \\
    Text Expert (Stage-2) & $56.7\%$ & $27.9\%$ & $70.9\%$ \\
    SPIRIT-LM \cite{nguyen2024spirit} (Llama2) & $36.9\%$ & $/$ & $/$ \\
    SPIRIT-LM \cite{nguyen2024spirit} (Llama3) & $53.5\%$ & $/$ & $/$ \\
    \coloredmidrule
    \rowcolor{lightblue}
    \textbf{TTS-Llama} & $27.2\%$ & $23.2\%$ & $24.8\%$ \\
    \rowcolor{lightblue}
    \textbf{MoLE-Llama (TTS+Text QA)} & $54.8\%$ & $26.1\%$ & $70.3\%$ \\
    \rowcolor{lightblue}
    \textbf{MoLE-Llama (Speech QA+Text QA)} & $53.9\%$ & $26.3\%$ & $71.1\%$ \\
    \coloredbottomrule
  \end{tabular}
  \label{table:text-qa}
  \end{center}
\vspace{-1.5em}
\end{table}

As evidenced in Table \ref{table:text-qa}, our proposed MoLE-Llama effectively mitigates the catastrophic forgetting issue observed in TTS-Llama. Both two versions of MoLE-Llama demonstrate strong performance on text QA tasks across three benchmark datasets. Moreover, while the SOTA method SPIRIT-LM~\cite{nguyen2024spirit} requires full fine-tuning of pretrained Llama models with 1-2 orders of magnitude more training data, MoLE-Llama achieves comparable performance on the MMLU benchmark using a purely late-fusion approach of PEFT fine-tuning.
\section{Concluding Remarks}
This work introduces two novel models. TTS-Llama demonstrates the effectiveness of enabling TTS generation via only PEFT fine-tuning of a text-based LLM. Building on TTS-Llama, MoLE-Llama further combines both text-QA and TTS capabilities using mixture of LoRA experts without catastrophic forgetting in either modality, highlighting the potential of late fusion in enabling text-speech multimodal LLMs. A natural extension of this work is to close the gap between the text-QA expert and the original text-only LLM. Additionally, the speech-output QA capability could be further enhanced instead of relying on the ``chain-of-modality'' technique.

\clearpage
\bibliographystyle{IEEEtran}
\bibliography{refs}

\end{document}